\documentclass[11pt]{article}

\usepackage{acl}
\usepackage{times}
\usepackage{latexsym}

\usepackage[T1]{fontenc}

\usepackage[utf8]{inputenc}

\usepackage{microtype}

\usepackage{inconsolata}

\usepackage{graphicx}

\usepackage{booktabs}
\usepackage{amsmath}

\usepackage[most]{tcolorbox}
\usepackage{tabularx}
\usepackage{xcolor}

\usepackage{listings}
\usepackage{algorithm}
\usepackage{algpseudocode}
\usepackage{amsmath}

\usepackage{booktabs}
\usepackage{multirow}
\usepackage{xcolor}
\usepackage{adjustbox}

\usepackage{booktabs}
\usepackage{multirow}
\usepackage{colortbl}
\usepackage{xcolor}
\usepackage{adjustbox}

\usepackage{multirow}

\usepackage{booktabs}
\usepackage{multirow}
\usepackage{adjustbox}
\usepackage{xcolor}

\usepackage{amsmath,amssymb,amsfonts}
\usepackage[most]{tcolorbox}
\tcbset{
  myboxstyle/.style={
    enhanced,
    breakable,
    colback=blue!5!white,
    colframe=blue!40!black,
    coltitle=black,
    fonttitle=\bfseries,
    title={#1},
    boxrule=0.4pt,
    arc=2pt,
    left=6pt,
    right=6pt,
    top=4pt,
    bottom=4pt
  }
}

\usepackage{multirow}
\usepackage[table]{xcolor}
\usepackage{graphicx}

\usepackage{amsmath}
\usepackage{amssymb}
\usepackage{algorithm}

\usepackage{placeins}

\usepackage{float}

\usepackage{needspace}

\newcommand{\algrule}[1][.4pt]{\par\vskip.5\baselineskip\hrule height #1\par\vskip.5\baselineskip}

\DeclareUnicodeCharacter{1F99C}{\textit{(parrot)}} 

\title{ZPD Detector: Data Selection via Capability–Difficulty Alignment for Large Language Models}

\author{
\textbf{Bo Yang$^{*}$, Yunkui Chen$^{*}$, Lanfei Feng, Yu Zhang, Shijian Li$^{\dagger}$} \\[3.5pt]
\textbf{State Key Laboratory of Brain–Machine Intelligence } \\[3.5pt]
\textbf{College of Computer Science and Technology, Zhejiang University, Hangzhou, China} \\[3.5pt]
\textbf{\{boyang30, 22351048, 22451116, 22421173, shijianli\}@zju.edu.cn}
}

\begin{document}
\maketitle

\begin{abstract}

As the cost of training large language models continues to increase and high-quality training data become increasingly scarce, selecting high-value samples or synthesizing effective training data under limited data budgets has emerged as a critical research problem. Most existing data selection methods rely on static criteria, such as difficulty, uncertainty, or heuristics, and fail to model the evolving relationship between the model and the data. Inspired by the educational theory of the \textbf{Zone of Proximal Development (ZPD)}, we propose \textbf{ZPD Detector}, a data selection framework that adopts a bidirectional perspective between models and data by explicitly modeling the alignment between sample difficulty and the model’s current capability. ZPD Detector integrates difficulty calibration, model capability estimation based on \textbf{Item Response Theory (IRT)}, and a capability--difficulty matching score to dynamically identify the most informative samples at each learning stage, improving data utilization efficiency; moreover, this dynamic matching strategy provides new insights into training strategy design.  All code and data will be released after our work be accepted to support reproducible research.

\end{abstract}

\section{Introduction}

\begin{figure}[t]
  \includegraphics[width=\columnwidth]{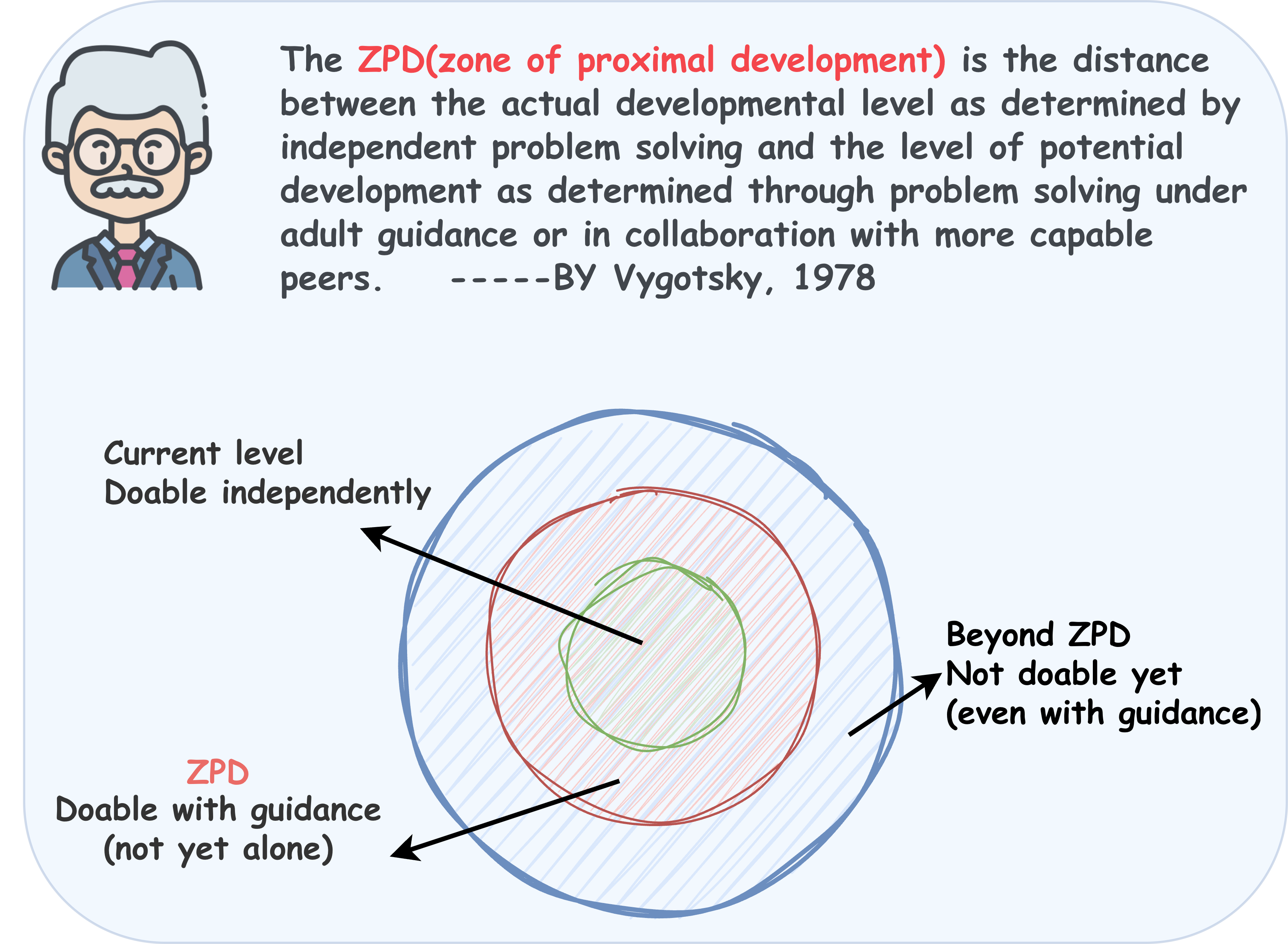}
  \caption{\textbf{Motivation of ZPD-based approach.} The most informative training samples lie between what a model can already solve and what is beyond its current capability.}
  \label{fig:experiments}
\end{figure}

\begin{figure*}[t]
\centering
\includegraphics[width=1\linewidth]{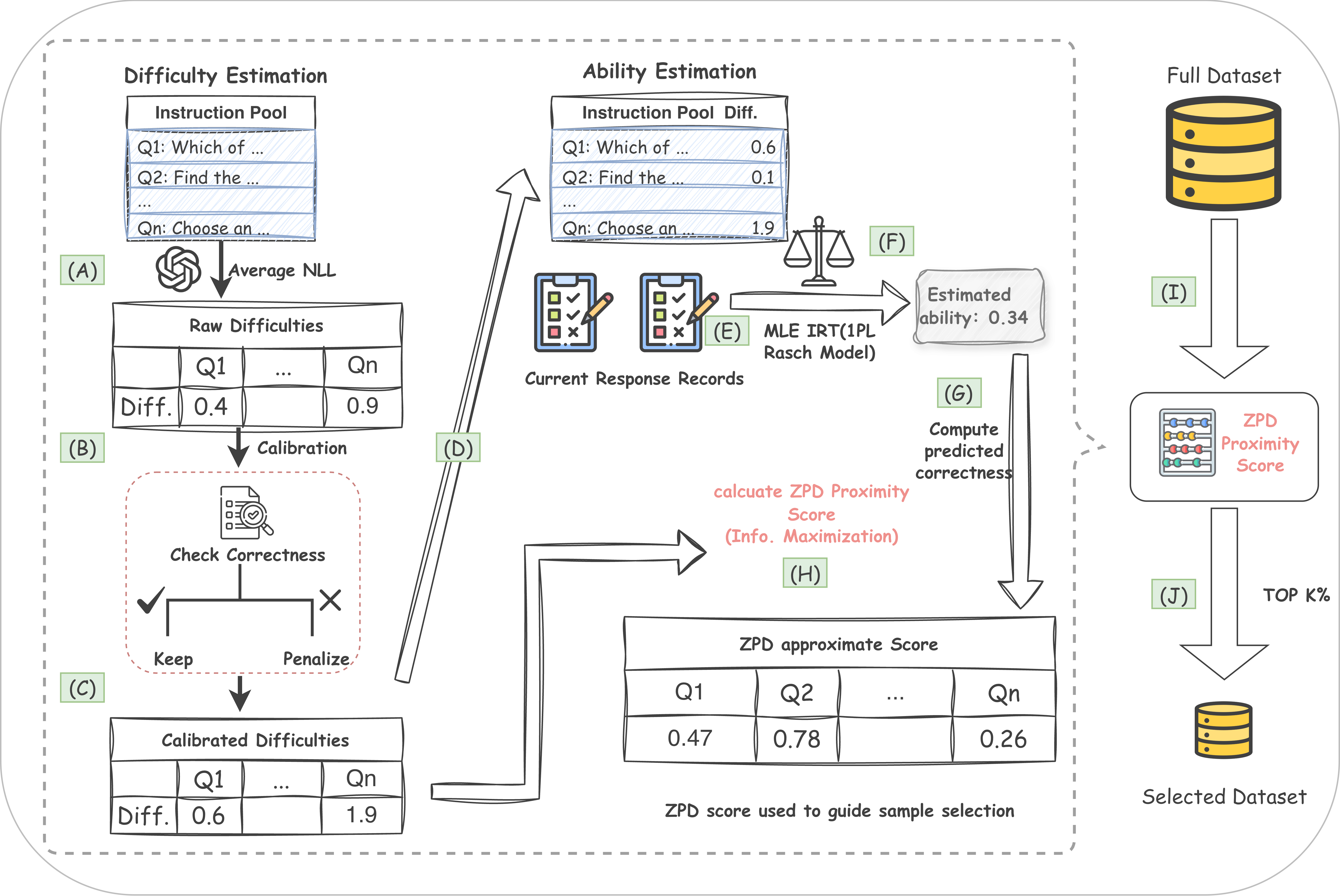}

\caption{
\textbf{Overview of the ZPD Detector pipeline.} 
\textbf{(A)} Estimate raw sample difficulty using average token-level NLL. 
\textbf{(B)} Adjust underestimated errors via model correctness feedback. 
\textbf{(C)} Normalize calibrated difficulty scores as item parameters. 
\textbf{(D)} Input difficulty into the Rasch model. 
\textbf{(E)} Use model correctness records as responses. 
\textbf{(F)} Estimate model ability via maximum likelihood. 
\textbf{(G)} Predict the probability of correctness for each sample. 
\textbf{(H)} Compute ZPDScores to reflect model uncertainty. 
\textbf{(I)} Rank samples by ZPDScores. 
\textbf{(J)} Select the top-$k\%$ most informative samples for fine-tuning.
}  
\label{fig:overview}
\end{figure*}

As large language models continue to scale, the cost of training and fine-tuning has increased rapidly; meanwhile, high-quality real-world training data are becoming increasingly scarce, especially in professional and domain-specific scenarios~\cite{kaplan2020scaling,hoffmann2022training}.
Under this setting, improving model performance under limited data budgets has emerged as a critical research problem~\cite{frankle2018lottery}.

Data selection has been widely regarded as an effective approach to alleviating this challenge~\cite{xie2023data,wang2018dataset,settles2009active}.
Most existing methods perform static selection based on sample difficulty, uncertainty, or heuristic rules, and implicitly assume that \textbf{the learning value of a sample is independent of the model state and can be treated as a static property}~\cite{xie2023data,settles2009active}.
However, in practice, the effectiveness of a sample strongly depends on the model’s current capability:
overly easy samples provide limited learning signals, while overly difficult ones may introduce noise or even hinder convergence~\cite{bengio2009curriculum,kumar2010self,zhang2016understanding,bender2021dangers}.
This observation suggests that data selection fundamentally involves a \emph{dynamic relationship} between model capability and sample difficulty~\cite{bengio2009curriculum}.

Inspired by the educational theory of the Zone of Proximal Development (ZPD)~\cite{vygotsky1978mind}, we revisit data selection from a \emph{bidirectional perspective} between models and data~\cite{kumar2010self}.
We argue that the most informative samples are those whose difficulty is well aligned with the model’s current capability, rather than those located at the extremes of the difficulty distribution.
Based on this insight, we propose \textbf{ZPD Detector}, a ZPD-centered data selection framework that explicitly models the relationship between sample difficulty and model capability through difficulty calibration~\cite{xie2023data,settles2009active}, model capability estimation based on Item Response Theory (IRT)~\cite{martinez2019item,lee2019estimating}, and capability--difficulty matching scores.
ZPD Detector dynamically identifies the most informative samples for the model at its current learning stage without modifying model architectures, making it plug-and-play and transferable~\cite{wang2018dataset}.
Beyond data selection, this capability--difficulty alignment principle can also be naturally incorporated into training strategy design, offering a new perspective for staged and adaptive training~\cite{matiisen2019teacher,wang2019dynamic}.

Our contributions are summarized as follows:

\begin{itemize}
    \item \textbf{A ZPD-centered perspective on data selection.}
    We shift data selection from evaluating the intrinsic quality of samples to modeling the alignment between sample difficulty and the model’s current capability, viewing model capacity and training data as a dynamically coupled system.

    \item \textbf{A plug-and-play and transferable ZPD Detector framework.}
    The proposed framework is modular, requires no modification to model architectures, and is agnostic to specific model designs, enabling effective transfer across models and datasets.

    \item \textbf{Systematic validation on real and synthetic benchmarks.}
    Experiments show that ZPD-selected subsets can match or even slightly exceed full-data training performance, while consistently outperforming existing data selection methods under the same data budget.

    \item \textbf{A generalizable training insight.}
    ZPD further reveals a general training principle that training signals should be dynamically aligned with model capability, which naturally extends to staged training strategy design.
\end{itemize}

\section{Related Work}

Data selection aims to improve data efficiency by ranking or filtering samples using loss, difficulty, gradient-based signals, or redundancy constraints~\citep{paul2021deep,killamsetty2021grad,mirzasoleiman2020coresets,wei2014submodular}. In large language models and instruction tuning, selection is used to reduce training and alignment cost via instruction ranking or large-scale filtering~\citep{li2024one,zhang2025active,jin2024optimizing,chen2025revisiting}, as well as model-aware criteria such as implicit rewards or gradient alignment~\citep{zhou2025davir,li2025learnalign}.

Most prior methods treat data utility as static, independent of model state, despite evidence that model capabilities evolve across architectures, scales, and training stages~\citep{toneva2018empirical,paul2021deep,xia2023training}. Recent work addresses this mismatch by incorporating capability or alignment awareness, including filtering overly difficult data~\citep{gao2025principled}, balancing data across capabilities~\citep{ming2025ideal}, and delegating data selection to smaller models~\citep{mekala2024smaller}.

Related studies further analyze fine-grained and predictable aspects of model capability, highlighting limitations of single-axis modeling and sensitivity to task formats~\citep{ge2025capability,suzgun2023challenging,ye2023predictable}. In contrast, we propose a symmetric, capability-aware selection framework inspired by the Zone of Proximal Development, explicitly aligning sample difficulty with the model’s current ability.

\section{Method}

Our goal is to enhance domain-specific instruction tuning under limited training budgets by selecting the most pedagogically effective subset of data. We propose \textbf{ZPD Detector}, a cognitively motivated framework inspired by the educational theory of the \textbf{ZPD}. According to this theory, optimal learning occurs when the learner engages with tasks that are not trivially easy nor impossibly hard, but lie near the boundary of their current competence—those that are \textit{doable with guidance}.

Translating this idea into the context of large language models (LLMs), we aim to identify samples that are neither fully mastered nor entirely unsolvable by the current model, but rather occupy a zone where the model exhibits uncertainty and hence potential for growth. To formalize this intuition, we adopt a probabilistic framework grounded in IRT, commonly used in psychometrics, and apply the 1PL Rasch model to jointly characterize model ability, sample difficulty, and prediction behavior.

Our framework comprises five stages: raw difficulty estimation, difficulty calibration, IRT-based difficulty modeling, ability estimation, and ZPD-based selection. We now describe each component in detail. The detailed algorithmic procedure is available in Appendix A.4.

\subsection{Step 1: Raw Difficulty Estimation}

In the absence of additional supervision, we require a scalable and generalizable proxy for sample difficulty. We adopt the average token-level negative log-likelihood (NLL) under teacher forcing as our primary measure. This metric reflects how well the current model fits the reference answer: the higher the NLL, the greater the model's difficulty with the sample.

Formally, given an instruction-output pair $(x_i, y_i)$, where $y_i = \{y_{i,1}, \dots, y_{i,L_i}\}$ denotes the reference response, we compute:

\[
\text{RawDiff}_i = -\frac{1}{L_i} \sum_{t=1}^{L_i} \log P(y_{i,t} \mid y_{i,<t}, x_i)
\]

A larger RawDiff indicates greater difficulty, as the model struggles to predict the reference answer. Conversely, low RawDiff values suggest the model has already mastered the underlying pattern, rendering the sample less informative for further training.

This metric is model-centered, task-agnostic, and compatible with diverse output formats. It can be computed offline in parallel before training, and serves as a stable foundation for subsequent modeling stages.

\subsection{Step 2: Difficulty Calibration}

While RawDiff provides an initial estimate of sample difficulty, it exhibits a systematic inconsistency: some samples answered incorrectly by the model still receive low NLL scores, leading to an underestimation of their effective difficulty. Such cases violate the expected monotonic relationship between difficulty and correctness and can bias subsequent ability modeling.

To correct this inconsistency, we apply a lightweight, error-aware calibration based on behavioral feedback. Let $r_i \in \{0, 1\}$ indicate whether sample $i$ is answered correctly, and let $\mu$ denote the mean RawDiff over the dataset. The calibrated difficulty is defined as:
\[
\tilde{d}_i = \text{RawDiff}_i + (1 - r_i)\cdot \max(0, \mu - \text{RawDiff}_i).
\]

This rule leaves correctly answered samples unchanged, while enforcing a lower-bound correction for confidently incorrect ones whose RawDiff falls below the dataset average. Importantly, the adjustment is local and minimal: samples that are already difficult receive no additional penalty, preventing distortion of the global difficulty distribution.

Rather than learning a full calibration function, this formulation enforces a consistency constraint between observed correctness and difficulty ordering. It is parameter-free, relies only on binary correctness signals, and remains robust under skewed or multi-modal difficulty distributions. The resulting calibrated difficulty $\tilde{d}_i$ provides a behaviorally consistent input for subsequent IRT-based modeling.

\subsection{Step 3: IRT-Based Difficulty Modeling}

Having calibrated the sample difficulties, we seek to unify the relationship between sample difficulty, model ability, and prediction outcome under a principled probabilistic framework. To this end, we adopt the \textbf{Item Response Theory (IRT)}—a foundational model from psychometrics—and apply its simplest yet most interpretable form, the one-parameter logistic (1PL) Rasch model. We adopt the Rasch (1PL) model as a minimal and stable abstraction of model capability, prioritizing robustness and identifiability over fine-grained parameterization.

In classical IRT, each "test item" (here, a training sample) is characterized by a difficulty parameter $b_i$, and each "test taker" (the model) is assigned a latent ability score $\theta$. The probability that the model correctly answers sample $i$ is given by:

\[
P_i = P(r_i = 1 \mid \theta, b_i) = \sigma(\theta - b_i) = \frac{1}{1 + e^{-(\theta - b_i)}}
\]

where $r_i \in \{0, 1\}$ denotes the correctness of the model's prediction, and $\sigma(\cdot)$ is the sigmoid function. When the model’s ability matches the item’s difficulty ($\theta = b_i$), the success probability is 50\%; the further $\theta$ exceeds $b_i$, the more likely the model is to answer correctly.

To integrate this into our task, we linearly normalize the calibrated difficulty $\tilde{d}_i$ into a reasonable value range to obtain the Rasch difficulty $b_i$. These difficulty values serve as a fixed scale for estimating model ability $\theta$ in the next step, and the predicted success probability $P_i$ becomes the foundation for our ZPD-based selection strategy.

The Rasch model offers dual benefits in our setting: it elevates difficulty modeling from heuristic scores to probabilistic reasoning, and its parsimonious structure ensures stable and interpretable performance, even under limited supervision or data availability—conditions common in domain-specific instruction tuning.

\subsection{Step 4: Ability Estimation}

With a fixed scale of Rasch-based sample difficulties $\{b_i\}$, the next step is to estimate the model’s current ability $\theta$ based on its observed prediction outcomes. Let $\mathcal{C}$ be the set of examples used for calibration, and let $r_i \in \{0,1\}$ indicate whether the model answered sample $i$ correctly. Under the Rasch model, the likelihood of these responses is:

\[
L(\theta) = \prod_{i \in \mathcal{C}} \sigma(\theta - b_i)^{r_i} \cdot (1 - \sigma(\theta - b_i))^{1 - r_i}
\]

Taking the logarithm yields the log-likelihood:

\[
\begin{aligned}
\log L(\theta) = \sum_{i \in \mathcal{C}} \big[ 
    & r_i \log \sigma(\theta - b_i) \\
    +\ & (1 - r_i) \log (1 - \sigma(\theta - b_i)) 
\big]
\end{aligned}
\]

We seek the value $\hat{\theta}$ that maximizes this log-likelihood. Since it is a smooth, unimodal function in one dimension, we apply a \textbf{bisection search} to efficiently locate the root of its derivative and obtain the maximum likelihood estimate. Details are available in Appendix A.5

The resulting ability estimate $\hat{\theta}$ offers a personalized summary of the model’s current competence on the given task. It serves as a key input for the final ZPD scoring mechanism, allowing us to determine which samples are most aligned with the model’s present learning frontier.

\subsection{Step 5: ZPD Scoring and Data Selection}

With the model ability $\hat{\theta}$ estimated and a Rasch-calibrated difficulty scale $\{b_i\}$ for each sample, we now seek to identify training examples that are most beneficial for further improvement. Inspired by Vygotsky’s Zone of Proximal Development (ZPD), we hypothesize that the most valuable samples lie in a region where the model is uncertain yet capable of learning—neither too easy (already mastered) nor too hard (currently unreachable).

To quantify this, we first compute the predicted probability of answering each sample correctly:

\[
p_i = \sigma(\hat{\theta} - b_i)
\]

To capture the model's uncertainty in a compact form, we define the \textbf{ZPDScore} as a symmetric function peaking at $p_i = 0.5$:

\[
\text{ZPDScore}_i = p_i (1 - p_i)
\]

This metric approximates the entropy curve while being simpler and more efficient to compute. It assigns the highest scores to samples closest to the decision boundary—where the model exhibits maximum uncertainty and thus greatest learning potential.

We sort all samples by ZPDScore in descending order and select the top-$\rho$ fraction, based on a predefined data budget ratio $\rho$:

\[
\mathcal{D}' = \text{TopK}_{i \in \mathcal{D}}(\text{ZPDScore}_i, \rho)
\]

This adaptive filtering strategy dynamically aligns with the model’s current state, implicitly implementing a form of curriculum learning. It promotes those samples the model is \textit{ready to learn}, rather than those it has already mastered or is unlikely to benefit from, resulting in significantly improved data efficiency across tasks and model scales.

\begin{table*}[t]
\centering
\small
\setlength{\tabcolsep}{3.6pt}
\renewcommand{\arraystretch}{1.12}

\resizebox{\textwidth}{!}{%
\begin{tabular}{l|cccc|cccc|cccc}
\hline
\textbf{Method}
& \multicolumn{4}{c|}{\textbf{MedQA}}
& \multicolumn{4}{c|}{\textbf{GSM8K}}
& \multicolumn{4}{c}{\textbf{AgriQA (Synthetic)}} \\
& 1\% & 5\% & 10\% & 15\%
& 1\% & 5\% & 10\% & 15\%
& 1\% & 5\% & 10\% & 15\% \\
\hline

\multicolumn{13}{l}{\textbf{LLaMA3-8B-Instruct}} \\
\textit{Zero-shot}
& \multicolumn{4}{c|}{53.18}
& \multicolumn{4}{c|}{32.37}
& \multicolumn{4}{c}{42.67} \\

+ Random
& 53.57 {\color{red}{$\downarrow$1.18}}
& 56.09 {\color{red}{$\downarrow$0.70}}
& 55.77 {\color{red}{$\downarrow$1.89}}
& 56.64 {\color{red}{$\downarrow$0.63}}
& 45.64 {\color{red}{$\downarrow$1.14}}
& 63.99 {\color{red}{$\downarrow$0.30}}
& 65.25 {\color{red}{$\downarrow$0.10}}
& 64.90 {\color{red}{$\downarrow$4.62}}
& 92.09 {\color{red}{$\downarrow$0.20}}
& 92.21 {\color{red}{$\downarrow$0.51}}
& 92.81 {\color{red}{$\downarrow$0.76}}
& 92.98 {\color{red}{$\downarrow$0.51}} \\

+ IFD
& 43.52 {\color{red}{$\downarrow$11.23}}
& 50.90 {\color{red}{$\downarrow$5.89}}
& 53.26 {\color{red}{$\downarrow$4.40}}
& 54.20 {\color{red}{$\downarrow$3.07}}
& 47.84 {\color{green}{$\uparrow$1.06}}
& 63.53 {\color{red}{$\downarrow$0.76}}
& 63.76 {\color{red}{$\downarrow$1.59}}
& 67.10 {\color{red}{$\downarrow$2.42}}
& 92.04 {\color{red}{$\downarrow$0.25}}
& 92.21 {\color{red}{$\downarrow$0.51}}
& 92.89 {\color{red}{$\downarrow$0.68}}
& 93.32 {\color{red}{$\downarrow$0.17}} \\

+ PPL
& 54.91 {\color{green}{$\uparrow$0.16}}
& 55.93 {\color{red}{$\downarrow$0.86}}
& 54.99 {\color{red}{$\downarrow$2.67}}
& 54.44 {\color{red}{$\downarrow$2.83}}
& 46.78 {\color{red}{$\downarrow$0.00}}
& 58.68 {\color{red}{$\downarrow$5.61}}
& 64.29 {\color{red}{$\downarrow$1.06}}
& 62.32 {\color{red}{$\downarrow$7.20}}
& 92.46 {\color{green}{$\uparrow$0.17}}
& 91.69 {\color{red}{$\downarrow$1.03}}
& 92.37 {\color{red}{$\downarrow$1.20}}
& 92.63 {\color{red}{$\downarrow$0.86}} \\

+ AlpaGasus
& 53.89 {\color{red}{$\downarrow$0.86}}
& 54.83 {\color{red}{$\downarrow$1.96}}
& 56.01 {\color{red}{$\downarrow$1.65}}
& 55.70 {\color{red}{$\downarrow$1.57}}
& 46.61 {\color{red}{$\downarrow$0.17}}
& 64.06 {\color{red}{$\downarrow$0.23}}
& 65.20 {\color{red}{$\downarrow$0.15}}
& 67.63 {\color{red}{$\downarrow$1.89}}
& 92.29 {\color{red}{$\downarrow$0.00}}
& 91.77 {\color{red}{$\downarrow$0.95}}
& 92.20 {\color{red}{$\downarrow$1.37}}
& 92.54 {\color{red}{$\downarrow$0.95}} \\

+ Data Whispererer
& 53.89 {\color{red}{$\downarrow$0.86}}
& 55.70 {\color{red}{$\downarrow$1.09}}
& 56.72 {\color{red}{$\downarrow$0.94}}
& 57.27 {\color{red}{$\downarrow$0.00}}
& 46.17 {\color{red}{$\downarrow$0.61}}
& 58.91 {\color{red}{$\downarrow$5.38}}
& 66.79 {\color{green}{$\uparrow$1.44}}
& 66.64 {\color{red}{$\downarrow$2.88}}
& 90.49 {\color{red}{$\downarrow$1.80}}
& 91.43 {\color{red}{$\downarrow$1.29}}
& 91.60 {\color{red}{$\downarrow$1.97}}
& 91.69 {\color{red}{$\downarrow$1.80}} \\

\rowcolor{cyan!12}
+ ZPD Detector
& \textbf{54.75} & \textbf{56.79} & \textbf{57.66} & \textbf{57.27}
& \textbf{46.78} & \textbf{64.29} & \textbf{65.35} & \textbf{69.52}
& \textbf{92.29} & \textbf{92.72} & \textbf{93.57} & \textbf{93.49} \\

+ Full Data
& \multicolumn{4}{c|}{58.05}
& \multicolumn{4}{c|}{69.29}
& \multicolumn{4}{c}{93.92} \\
\hline

\multicolumn{13}{l}{\textbf{Qwen3-8B}} \\
\textit{Zero-shot}
& \multicolumn{4}{c|}{55.15}
& \multicolumn{4}{c|}{41.62}
& \multicolumn{4}{c}{90.32} \\

+ Random
& 56.25 {\color{red}{$\downarrow$2.67}}
& 59.33 {\color{red}{$\downarrow$0.53}}
& 60.01 {\color{red}{$\downarrow$0.56}}
& 59.78 {\color{red}{$\downarrow$0.86}}
& 59.74 {\color{red}{$\downarrow$0.00}}
& 80.41 {\color{red}{$\downarrow$0.94}}
& 89.84 {\color{red}{$\downarrow$1.14}}
& 88.32 {\color{red}{$\downarrow$0.61}}
& 91.00 {\color{red}{$\downarrow$0.60}}
& 91.43 {\color{red}{$\downarrow$0.69}}
& 91.77 {\color{red}{$\downarrow$0.26}}
& 91.77 {\color{red}{$\downarrow$0.43}} \\

+ IFD
& 56.25 {\color{red}{$\downarrow$2.67}}
& 58.84 {\color{red}{$\downarrow$1.02}}
& 59.47 {\color{red}{$\downarrow$1.10}}
& 59.07 {\color{red}{$\downarrow$1.57}}
& 60.12 {\color{green}{$\uparrow$0.38}}
& 79.73 {\color{red}{$\downarrow$1.62}}
& 90.14 {\color{red}{$\downarrow$0.84}}
& 87.87 {\color{red}{$\downarrow$1.06}}
& 91.17 {\color{red}{$\downarrow$0.43}}
& 91.52 {\color{red}{$\downarrow$0.60}}
& 91.77 {\color{red}{$\downarrow$0.26}}
& 91.95 {\color{red}{$\downarrow$0.25}} \\

+ PPL
& 48.49 {\color{red}{$\downarrow$10.43}}
& 57.19 {\color{red}{$\downarrow$2.67}}
& 55.77 {\color{red}{$\downarrow$4.80}}
& 55.15 {\color{red}{$\downarrow$5.49}}
& 59.67 {\color{red}{$\downarrow$0.07}}
& 79.30 {\color{red}{$\downarrow$2.05}}
& 87.49 {\color{red}{$\downarrow$3.49}}
& 88.55 {\color{red}{$\downarrow$0.38}}
& 91.09 {\color{red}{$\downarrow$0.51}}
& 91.52 {\color{red}{$\downarrow$0.60}}
& 91.77 {\color{red}{$\downarrow$0.26}}
& 92.12 {\color{red}{$\downarrow$0.08}} \\

+ AlpaGasus
& 56.87 {\color{red}{$\downarrow$2.05}}
& 58.99 {\color{red}{$\downarrow$0.87}}
& 58.76 {\color{red}{$\downarrow$1.81}}
& 59.07 {\color{red}{$\downarrow$1.57}}
& 58.76 {\color{red}{$\downarrow$0.98}}
& 80.67 {\color{red}{$\downarrow$0.68}}
& 90.22 {\color{red}{$\downarrow$0.76}}
& 88.30 {\color{red}{$\downarrow$0.63}}
& 91.09 {\color{red}{$\downarrow$0.51}}
& 91.52 {\color{red}{$\downarrow$0.60}}
& 91.52 {\color{red}{$\downarrow$0.51}}
& 91.52 {\color{red}{$\downarrow$0.68}} \\

+ Data Whispererer
& 56.40 {\color{red}{$\downarrow$2.52}}
& 59.31 {\color{red}{$\downarrow$0.55}}
& 60.49 {\color{red}{$\downarrow$0.08}}
& 60.17 {\color{red}{$\downarrow$0.47}}
& 60.88 {\color{green}{$\uparrow$1.14}}
& 80.44 {\color{red}{$\downarrow$0.91}}
& 86.43 {\color{red}{$\downarrow$4.55}}
& 87.49 {\color{red}{$\downarrow$1.44}}
& 90.49 {\color{red}{$\downarrow$1.11}}
& 91.43 {\color{red}{$\downarrow$0.69}}
& 91.60 {\color{red}{$\downarrow$0.43}}
& 91.69 {\color{red}{$\downarrow$0.51}} \\

\rowcolor{cyan!12}
+ ZPD Detector
& \textbf{58.92} & \textbf{59.86} & \textbf{60.57} & \textbf{60.64}
& \textbf{59.74} & \textbf{81.35} & \textbf{90.98} & \textbf{88.93}
& \textbf{91.60} & \textbf{92.12} & \textbf{92.03} & \textbf{92.20} \\

+ Full Data
& \multicolumn{4}{c|}{60.88}
& \multicolumn{4}{c|}{90.43}
& \multicolumn{4}{c}{93.83} \\
\hline

\multicolumn{13}{l}{\textbf{Mistral-7B}} \\
\textit{Zero-shot}
& \multicolumn{4}{c|}{40.30}
& \multicolumn{4}{c|}{15.31}
& \multicolumn{4}{c}{86.03} \\

+ Random
& 41.08 {\color{red}{$\downarrow$1.34}}
& 45.25 {\color{red}{$\downarrow$0.45}}
& 44.85 {\color{red}{$\downarrow$1.30}}
& 45.88 {\color{red}{$\downarrow$0.80}}
& 29.42 {\color{red}{$\downarrow$6.29}}
& 41.47 {\color{red}{$\downarrow$1.67}}
& 45.87 {\color{red}{$\downarrow$2.42}}
& 46.70 {\color{red}{$\downarrow$4.78}}
& 87.83 {\color{red}{$\downarrow$0.86}}
& 89.72 {\color{red}{$\downarrow$0.25}}
& 90.57 {\color{red}{$\downarrow$0.69}}
& 91.07 {\color{red}{$\downarrow$0.79}} \\

+ IFD
& 33.94 {\color{red}{$\downarrow$8.48}}
& 21.45 {\color{red}{$\downarrow$24.25}}
& 21.20 {\color{red}{$\downarrow$24.95}}
& 21.87 {\color{red}{$\downarrow$24.81}}
& 28.43 {\color{red}{$\downarrow$7.28}}
& 40.18 {\color{red}{$\downarrow$2.96}}
& 44.66 {\color{red}{$\downarrow$3.63}}
& 47.38 {\color{red}{$\downarrow$4.10}}
& 85.86 {\color{red}{$\downarrow$2.83}}
& 87.08 {\color{red}{$\downarrow$2.89}}
& 88.35 {\color{red}{$\downarrow$2.91}}
& 91.09 {\color{red}{$\downarrow$0.77}} \\

+ PPL
& 40.77 {\color{red}{$\downarrow$1.65}}
& 21.60 {\color{red}{$\downarrow$24.10}}
& 18.70 {\color{red}{$\downarrow$27.45}}
& 22.31 {\color{red}{$\downarrow$24.37}}
& 16.15 {\color{red}{$\downarrow$19.56}}
& 33.21 {\color{red}{$\downarrow$9.93}}
& 36.85 {\color{red}{$\downarrow$11.44}}
& 39.12 {\color{red}{$\downarrow$12.36}}
& 86.20 {\color{red}{$\downarrow$2.49}}
& 88.85 {\color{red}{$\downarrow$1.12}}
& 89.29 {\color{red}{$\downarrow$1.97}}
& 90.49 {\color{red}{$\downarrow$1.37}} \\

+ AlpaGasus
& 42.58 {\color{green}{$\uparrow$0.16}}
& 42.83 {\color{red}{$\downarrow$2.87}}
& 45.09 {\color{red}{$\downarrow$1.06}}
& 45.25 {\color{red}{$\downarrow$1.43}}
& 30.93 {\color{red}{$\downarrow$4.78}}
& 38.51 {\color{red}{$\downarrow$4.63}}
& 45.87 {\color{red}{$\downarrow$2.42}}
& 47.54 {\color{red}{$\downarrow$3.94}}
& 87.40 {\color{red}{$\downarrow$1.29}}
& 89.80 {\color{red}{$\downarrow$0.17}}
& 90.32 {\color{red}{$\downarrow$0.94}}
& 90.23 {\color{red}{$\downarrow$1.63}} \\

+ Data Whispererer
& 38.33 {\color{red}{$\downarrow$4.09}}
& 44.46 {\color{red}{$\downarrow$1.24}}
& 45.90 {\color{red}{$\downarrow$0.25}}
& 46.21 {\color{red}{$\downarrow$0.47}}
& 27.45 {\color{red}{$\downarrow$8.26}}
& 41.47 {\color{red}{$\downarrow$1.67}}
& 44.66 {\color{red}{$\downarrow$3.63}}
& 45.72 {\color{red}{$\downarrow$5.76}}
& 87.57 {\color{red}{$\downarrow$1.12}}
& 89.03 {\color{red}{$\downarrow$0.94}}
& 89.63 {\color{red}{$\downarrow$1.63}}
& 90.57 {\color{red}{$\downarrow$1.29}} \\

\rowcolor{cyan!12}
+ ZPD Detector
& \textbf{42.42} & \textbf{45.70} & \textbf{46.15} & \textbf{46.68}
& \textbf{35.71} & \textbf{43.14} & \textbf{48.29} & \textbf{51.48}
& \textbf{88.69} & \textbf{89.97} & \textbf{91.26} & \textbf{91.86} \\

+ Full Data
& \multicolumn{4}{c|}{50.90}
& \multicolumn{4}{c|}{57.01}
& \multicolumn{4}{c}{92.66} \\
\hline

\end{tabular}%
}

\caption{
Comparison of data selection strategies on MedQA, GSM8K, and AgriQA (Synthetic) under varying data budgets.
Zero-shot and full-data results are reported as lower- and upper-bound references, respectively.
Performance differences ($\downarrow$ / $\uparrow$) are computed relative to the ZPD Detector at the same budget.
}
\label{tab:zpd_comparison_full}
\end{table*}

\section{Results}

\subsection{Experimental Setup}
\label{subsec:experimental_setup}

\paragraph{Datasets.}
We conduct experiments on three question answering datasets, each corresponding to a different type of downstream task.
Specifically, MedQA\citep{jin2021disease} represents knowledge-intensive question answering in professional domains,
GSM8K \citep{cobbe2021training} focuses on logical and mathematical reasoning with multi-step solutions,
and AgriQA (Synthetic) is used to evaluate domain-specific factual question answering in the agricultural domain. The hyperparameter settings and dataset details are available in Appendix A.3.

MedQA and GSM8K are mature public benchmarks that reflect the behavior of data selection methods under real-world task distributions.
In contrast, AgriQA is constructed in a synthetic manner, with the primary goal of providing a controllable evaluation environment in terms of sample difficulty structure,
thereby enabling a more direct analysis of ZPD under different difficulty distributions.
In addition, this setting allows us to examine the generalization of ZPD to synthetic data scenarios.
Details of the data construction and quality control for AgriQA are provided in Appendix~A.7.

\paragraph{Models.}
Across all tasks, we evaluate three widely used open-source instruction-tuned large language models with comparable parameter scales but different reasoning and generalization characteristics,
namely LLaMA3-8B-Instruct \citep{dubey2024llama},
Qwen3-8B \citep{yang2025qwen3},
and Mistral-7B-Instruct \citep{jiang2023mistral}.

\paragraph{Baselines.}
We compare the proposed ZPD Detector with several representative data selection methods,
including (i) Random Selection, which samples subsets uniformly at random;
(ii) PPL-based Selection \citep{yin2024compute}, which ranks samples according to language model perplexity;
(iii) IFD \citep{li2024quantity} and AlpaGasus \citep{chen2023alpagasus}, which represent information-driven or heuristic-based data selection strategies based on sample difficulty;
and (iv) Data Whisperererer \citep{wang2025data}, a recent training-free data selection method based on model feedback.

In addition, we report Zero-shot performance (without any training) as a lower-bound reference,
and Full Data training (using the entire dataset) as an upper-bound reference.

\paragraph{Evaluation.}
Model performance is evaluated using Accuracy on MedQA and AgriQA,
and Exact Match (EM) on GSM8K by comparing the final generated answers with the ground-truth solutions.
Task-Specific prompt design is available in Appendix A.6.

\subsection{Main Results and Key Observations}
\label{subsec:main_results}

Table~\ref{tab:zpd_comparison_full} reports the comparative results of ZPD Detector against multiple data selection baselines on MedQA, GSM8K, and AgriQA (Synthetic), across three instruction-tuned large language models and varying data budgets. Overall, ZPD demonstrates consistent advantages across different tasks, models, and data regimes.

\paragraph{Effectiveness under Limited Data Budgets.}
On the real-world datasets MedQA and GSM8K, ZPD enables models to achieve performance comparable to, and in some cases exceeding, full-data fine-tuning while using substantially fewer training samples. 
For instance, on GSM8K with Qwen3-8B, fine-tuning on only 10\% of the training data selected by ZPD attains an Exact Match score of 90.98, which is slightly higher than the 90.43 achieved by training on the entire dataset.
Similar trends are observed for LLaMA3-8B-Instruct and Mistral-7B, indicating that ZPD effectively preserves the most informative samples under constrained data budgets.

These results suggest that, amid the rapidly increasing cost of large-scale model training and fine-tuning, principled data selection offers an alternative to scaling data volume. ZPD provides a practical pathway to improving data efficiency, allowing models to acquire critical knowledge and reasoning patterns with substantially reduced training data.

\paragraph{Consistent Gains over Baselines under the Same Budget.}
When compared under identical data budgets, ZPD Detector consistently outperforms random sampling and existing data selection methods across almost all settings.
As shown in Table~\ref{tab:zpd_comparison_full}, ZPD achieves the best or tied-best performance on MedQA and GSM8K across data ratios ranging from 1\% to 15\%, and this advantage remains stable across different backbone models.

Compared to perplexity-based selection (PPL) and heuristic difficulty-driven approaches (IFD and AlpaGasus), ZPD exhibits greater robustness and consistency. This observation indicates that explicitly modeling the alignment between sample difficulty and model capability yields more reliable performance gains than relying on single-factor heuristics.

\paragraph{Insights from Synthetic Data.}
On AgriQA (Synthetic), ZPD maintains clear and consistent advantages across all evaluated models and data proportions.
Unlike real-world datasets, AgriQA features a controllable difficulty structure, which allows the behavior of different data selection strategies to be examined more directly.
Under this setting, ZPD consistently achieves the best performance and approaches the upper bound established by full-data training.

These results highlight that ZPD does not depend on specific natural data distributions. More importantly, in the context where high-quality real-world data in vertical domains is becoming increasingly scarce and synthetic data is playing a growing role in model training, ZPD offers a meaningful insight into how to identify learning-effective samples in synthetic data regimes.

\begin{table*}[t]
\centering
\small
\setlength{\tabcolsep}{4pt}
\renewcommand{\arraystretch}{1.2}

\resizebox{\textwidth}{!}{%
\begin{tabular}{l|c|c|c|c|c|c}
\hline
\textbf{Dataset} & \textbf{Ratio}
& \textbf{Random}
& \textbf{w/o Calibration}
& \textbf{w/o Ability Est.}
& \textbf{w/o ZPDScore}
& \textbf{Our Method} \\
\hline

\multirow{6}{*}{\textbf{MedQA}}
& Zero-shot & \multicolumn{5}{c}{55.15} \\

& 1\%
& 56.25
& 58.13 {\color{red}{$\downarrow$0.79}}
& 50.35 {\color{red}{$\downarrow$8.57}}
& 50.35 {\color{red}{$\downarrow$8.57}}
& \cellcolor{cyan!12}\textbf{58.92} \\

& 5\%
& 59.33
& 59.39 {\color{red}{$\downarrow$0.47}}
& 57.11 {\color{red}{$\downarrow$2.75}}
& 57.27 {\color{red}{$\downarrow$2.59}}
& \cellcolor{cyan!12}\textbf{59.86} \\

& 10\%
& 60.01
& 60.57 {\color{red}{$\downarrow$0.00}}
& 55.22 {\color{red}{$\downarrow$5.35}}
& 52.24 {\color{red}{$\downarrow$8.33}}
& \cellcolor{cyan!12}\textbf{60.57} \\

& 15\%
& 59.78
& 59.94 {\color{red}{$\downarrow$0.70}}
& 55.30 {\color{red}{$\downarrow$5.34}}
& 55.46 {\color{red}{$\downarrow$5.18}}
& \cellcolor{cyan!12}\textbf{60.64} \\
\hline

\multirow{6}{*}{\textbf{GSM8K}}
& Zero-shot & \multicolumn{5}{c}{41.62} \\

& 1\%
& 59.74
& 60.42 {\color{red}{$\downarrow$0.68}}
& 59.67 {\color{red}{$\downarrow$0.07}}
& 59.58 {\color{red}{$\downarrow$0.16}}
& \cellcolor{cyan!12}\textbf{59.74} \\

& 5\%
& 80.41
& 79.15 {\color{red}{$\downarrow$2.20}}
& 78.54 {\color{red}{$\downarrow$2.81}}
& 78.77 {\color{red}{$\downarrow$2.58}}
& \cellcolor{cyan!12}\textbf{81.35} \\

& 10\%
& 89.84
& 87.79 {\color{red}{$\downarrow$3.19}}
& 87.87 {\color{red}{$\downarrow$3.11}}
& 87.64 {\color{red}{$\downarrow$3.34}}
& \cellcolor{cyan!12}\textbf{90.98} \\

& 15\%
& 88.32
& 88.54 {\color{red}{$\downarrow$0.39}}
& 88.39 {\color{red}{$\downarrow$0.54}}
& 88.78 {\color{red}{$\downarrow$0.15}}
& \cellcolor{cyan!12}\textbf{88.93} \\
\hline

\multirow{6}{*}{\textbf{AgriQA}}
& Zero-shot & \multicolumn{5}{c}{90.32} \\

& 1\%
& 91.00
& 90.57 {\color{red}{$\downarrow$1.03}}
& 90.49 {\color{red}{$\downarrow$1.11}}
& 90.49 {\color{red}{$\downarrow$1.11}}
& \cellcolor{cyan!12}\textbf{91.60} \\

& 5\%
& 91.43
& 91.69 {\color{red}{$\downarrow$0.43}}
& 90.92 {\color{red}{$\downarrow$1.20}}
& 90.92 {\color{red}{$\downarrow$1.20}}
& \cellcolor{cyan!12}\textbf{92.12} \\

& 10\%
& 91.77
& 91.69 {\color{red}{$\downarrow$0.34}}
& 91.35 {\color{red}{$\downarrow$0.68}}
& 91.35 {\color{red}{$\downarrow$0.68}}
& \cellcolor{cyan!12}\textbf{92.03} \\

& 15\%
& 91.77
& 92.20 {\color{red}{$\downarrow$0.00}}
& 91.95 {\color{red}{$\downarrow$0.25}}
& 91.95 {\color{red}{$\downarrow$0.25}}
& \cellcolor{cyan!12}\textbf{92.20} \\
\hline
\end{tabular}
}
\caption{Ablation study of the proposed ZPD-based data selection framework on Qwen3-8B.
Zero-shot results are reported separately to indicate the inherent capability of the backbone. W/O ZPDScore ranks samples by calibrated difficulty only, without ability estimation.}
\label{tab:zpd_ablation}
\end{table*}

\begin{table}[t]
\centering
\small
\setlength{\tabcolsep}{6pt}
\renewcommand{\arraystretch}{1.2}

\begin{tabular}{l c >{\columncolor{cyan!12}}c c}
\hline
\textbf{Dataset} & \textbf{EASY (10\%)} & \textbf{ZPD (10\%)} & \textbf{HARD (10\%)} \\
\hline
MedQA  & 58.84 & \textbf{60.57} & 52.24 \\
GSM8K  & 81.12 & \textbf{90.98} & 87.64 \\
AgriQA & 90.66 & \textbf{92.03} & 91.35 \\
\hline
\end{tabular}

\caption{
Ablation study comparing EASY-only, ZPD-selected, and HARD-only subsets under a fixed 10\% data budget.
The ZPD partition (highlighted) consistently outperforms both EASY and HARD selections across datasets,
demonstrating the benefit of aligning sample difficulty with model capability.
}
\label{tab:zpd_easy_hard}
\end{table}

\subsection{Ablation Studies}
\label{subsec:ablation}

We conduct ablation studies from two complementary perspectives to analyze the key design choices of ZPD Detector: 
(i) a component-wise analysis of the internal modules, and 
(ii) a theory-oriented validation of the Zone of Proximal Development (ZPD) hypothesis.
The results are summarized in Table~2 and Table~3, respectively.

\paragraph{Component-wise Ablation.}
Table~2 presents a component-wise ablation of ZPD Detector by individually removing its core modules, including difficulty calibration, model ability estimation, and the final ZPD scoring function.
Across different datasets and data budgets, removing any single component consistently leads to noticeable performance degradation, with the effect being more pronounced under low-resource settings.

Specifically, removing difficulty calibration weakens the reliability of difficulty estimation, making the selection process more sensitive to model bias and noisy samples.
Removing ability estimation causes the selection strategy to ignore the model’s current learning stage, resulting in a mismatch between selected samples and model capability.
Among all variants, removing the ZPD scoring function leads to the most substantial performance drop, indicating that difficulty or ability signals alone are insufficient for effective data selection.

\begin{figure}[t]
\centering
\includegraphics[width=1\linewidth]{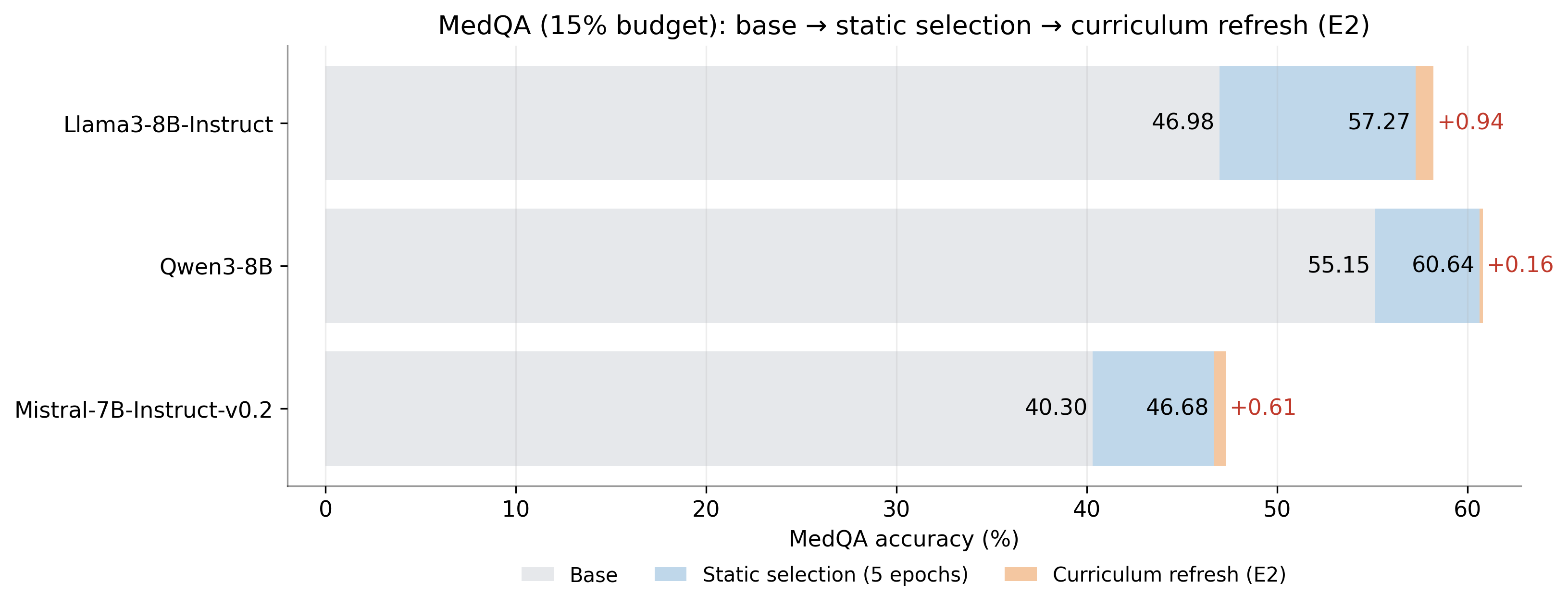}

\caption{
MedQA performance under a 15\% data budget comparing base training, static ZPD selection, and curriculum refresh (E2), demonstrating the benefit of capability-aware sample re-alignment during training.
}
\label{fig:train}
\end{figure}

\paragraph{ZPD Theory-oriented Ablation.}
Table~3 further evaluates ZPD from a theoretical perspective by directly testing the Zone of Proximal Development hypothesis.
Under a fixed 10\% data budget, we compare three selection strategies: selecting only low-difficulty samples (EASY-only), only high-difficulty samples (HARD-only), and samples located near the model’s current capability level (ZPD-selected).

The results show that ZPD-selected subsets consistently achieve the best performance across all datasets, significantly outperforming both EASY-only and HARD-only settings.
On MedQA and GSM8K, EASY-only selection yields stable but limited learning signals, while HARD-only selection often introduces overly difficult or noisy samples that hinder effective learning.
In contrast, ZPD-selected samples strike a better balance between stability and challenge, leading to higher training efficiency.

These findings provide empirical support for the core ZPD hypothesis: the most informative training samples are neither the easiest nor the hardest, but those whose difficulty is well aligned with the model’s current capability.

\begin{figure}[t]
  \includegraphics[width=\columnwidth]{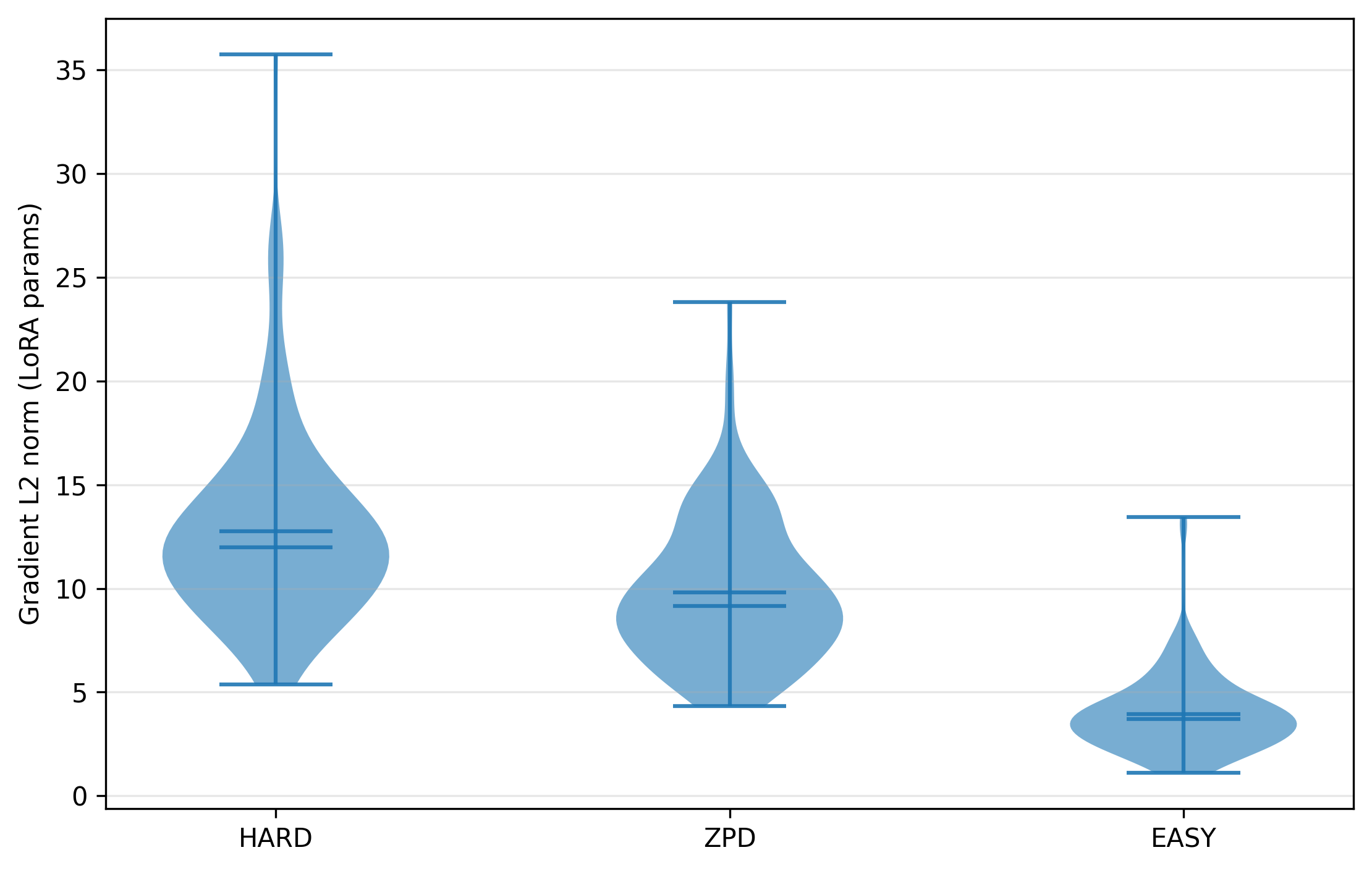}
  \caption{Gradient norm distributions of samples from different difficulty regions on Qwen3-8B. Each bucket contains an equal number of samples. Gradient norms are computed on LoRA-adapted layers in the transformer blocks.}
  \label{fig:Grad2}
\end{figure}

\section{Discussion and Insights}

\paragraph{Capability-adaptive behavior of ZPD (Fig.~\ref{fig:Grad1} is available in Appendix A.1.)} Difficulty distributions across different backbone models show that the target difficulty range selected by ZPD systematically shifts with model capability, rather than remaining fixed at an absolute difficulty level.
This indicates that ZPD explicitly captures the relative relationship between sample difficulty and model capability, validating the core assumption that a sample’s learning value depends on the model’s current capability state rather than being a static property.

\paragraph{Stable and effective learning signals within the ZPD region (Fig.~\ref{fig:Grad2}).}
From the perspective of optimization dynamics, Easy samples produce relatively weak gradients, while Hard samples exhibit larger and more dispersed gradient magnitudes, which may introduce training instability.
In contrast, samples within the ZPD region achieve a better balance between gradient strength and stability, suggesting that ZPD focuses on samples that provide efficient and reliable learning signals for the current model.

\paragraph{A unified view from data selection to training strategy (Fig.~\ref{fig:train}).}
Further experiments demonstrate that the ZPD principle extends beyond static data selection and can also guide sample scheduling during training.
By periodically re-aligning the difficulty distribution of training samples with the model’s evolving capability (curriculum refresh), additional performance gains can be achieved.
These results suggest that ZPD offers a unified capability-aware perspective that supports the joint design of data selection and training strategies.

\section{Conclusion}

We propose ZPD Detector, a capability-aware data selection method that dynamically aligns training data with model capability. Under limited data budgets, ZPD Detector achieves near or superior full-data performance using substantially fewer samples. Further analysis reveals that ZPD selection produces more stable and effective training signals, providing insights into efficient model training.
\section*{Limitations}

\paragraph{Single-dimensional ability modeling.}
Our formulation models model capability using a single latent ability with unit discrimination under the Rasch (1PL) framework. 
While this design favors stability and identifiability, it may be insufficient for tasks involving heterogeneous skills or multiple capability dimensions. 
Extending ZPD to multi-dimensional or higher-order IRT models (e.g., 2PL) is a promising direction for future work.

\paragraph{Calibration and prompting dependence.}
Our difficulty calibration relies on binary correctness signals and enforces a minimal consistency constraint rather than learning a full mapping from loss to correctness. 
This design improves robustness under limited data and skewed difficulty distributions, but it also assumes fixed prompting and decoding settings for ability estimation. 
For tasks that are sensitive to generation formats or prompt variations, such assumptions may affect the estimated capability and subsequent data selection.

\paragraph{Selection overhead.}
The selection procedure requires additional forward passes to compute NLL and correctness signals. 
In our setup, this overhead is comparable to perplexity-based filtering and can be amortized over training, but it may become non-negligible for extremely large-scale corpora.

\paragraph{Limitations and Potential Risks.}
ZPD Detector depends on estimated model capability and sample difficulty, which may be noisy or imperfect under domain shift. Inaccurate difficulty estimation can lead to suboptimal data selection in some settings. The method is intended to improve training efficiency rather than guarantee optimal performance and should be validated when applied to new tasks or models.

\bibliography{custom}

@article{paul2021deep,
  title={Deep learning on a data diet: Finding important examples early in training},
  author={Paul, Mansheej and Ganguli, Surya and Dziugaite, Gintare Karolina},
  journal={Advances in neural information processing systems},
  volume={34},
  pages={20596--20607},
  year={2021}
}

@inproceedings{killamsetty2021grad,
  title={Grad-match: Gradient matching based data subset selection for efficient deep model training},
  author={Killamsetty, Krishnateja and Durga, Sivasubramanian and Ramakrishnan, Ganesh and De, Abir and Iyer, Rishabh},
  booktitle={International Conference on Machine Learning},
  pages={5464--5474},
  year={2021},
  organization={PMLR}
}

@inproceedings{mirzasoleiman2020coresets,
  title={Coresets for data-efficient training of machine learning models},
  author={Mirzasoleiman, Baharan and Bilmes, Jeff and Leskovec, Jure},
  booktitle={International Conference on Machine Learning},
  pages={6950--6960},
  year={2020},
  organization={PMLR}
}

@inproceedings{wei2014submodular,
  title={Submodular subset selection for large-scale speech training data},
  author={Wei, Kai and Liu, Yuzong and Kirchhoff, Katrin and Bartels, Chris and Bilmes, Jeff},
  booktitle={2014 IEEE International Conference on Acoustics, Speech and Signal Processing (ICASSP)},
  pages={3311--3315},
  year={2014},
  organization={IEEE}
}

@inproceedings{li2024one,
  title={One-shot learning as instruction data prospector for large language models},
  author={Li, Yunshui and Hui, Binyuan and Xia, Xiaobo and Yang, Jiaxi and Yang, Min and Zhang, Lei and Si, Shuzheng and Chen, Ling-Hao and Liu, Junhao and Liu, Tongliang and others},
  booktitle={Proceedings of the 62nd Annual Meeting of the Association for Computational Linguistics (Volume 1: Long Papers)},
  pages={4586--4601},
  year={2024}
}

@inproceedings{zhang2025active,
  title={Active Instruction Tuning for Large Language Models with Reference-Free Instruction Selection},
  author={Zhang, Qiyuan and Chen, Jiehao and Ma, Chen},
  booktitle={Pacific-Asia Conference on Knowledge Discovery and Data Mining},
  pages={16--28},
  year={2025},
  organization={Springer}
}

@article{jin2024optimizing,
  title={Optimizing dataset creation: A general purpose data filtering system for training large language models},
  author={Jin, Sigo and Wang, Yanbing and Liu, Shan and Zhang, Yue and Gu, Wei},
  year={2024}
}

@inproceedings{chen2025revisiting,
  title={Revisiting scaling laws for language models: The role of data quality and training strategies},
  author={Chen, Zhengyu and Wang, Siqi and Xiao, Teng and Wang, Yudong and Chen, Shiqi and Cai, Xunliang and He, Junxian and Wang, Jingang},
  booktitle={Proceedings of the 63rd Annual Meeting of the Association for Computational Linguistics (Volume 1: Long Papers)},
  pages={23881--23899},
  year={2025}
}

@article{toneva2018empirical,
  title={An empirical study of example forgetting during deep neural network learning},
  author={Toneva, Mariya and Sordoni, Alessandro and Combes, Remi Tachet des and Trischler, Adam and Bengio, Yoshua and Gordon, Geoffrey J},
  journal={arXiv preprint arXiv:1812.05159},
  year={2018}
}

@article{kaplan2020scaling,
  title={Scaling laws for neural language models},
  author={Kaplan, Jared and McCandlish, Sam and Henighan, Tom and Brown, Tom B and Chess, Benjamin and Child, Rewon and Gray, Scott and Radford, Alec and Wu, Jeffrey and Amodei, Dario},
  journal={arXiv preprint arXiv:2001.08361},
  year={2020}
}

@article{hoffmann2022training,
  title={Training compute-optimal large language models},
  author={Hoffmann, Jordan and Borgeaud, Sebastian and Mensch, Arthur and Buchatskaya, Elena and Cai, Trevor and Rutherford, Eliza and Casas, Diego de Las and Hendricks, Lisa Anne and Welbl, Johannes and Clark, Aidan and others},
  journal={arXiv preprint arXiv:2203.15556},
  year={2022}
}

@article{frankle2018lottery,
  title={The lottery ticket hypothesis: Finding sparse, trainable neural networks},
  author={Frankle, Jonathan and Carbin, Michael},
  journal={arXiv preprint arXiv:1803.03635},
  year={2018}
}

@article{xie2023data,
  title={Data selection for language models via importance resampling},
  author={Xie, Sang Michael and Santurkar, Shibani and Ma, Tengyu and Liang, Percy S},
  journal={Advances in Neural Information Processing Systems},
  volume={36},
  pages={34201--34227},
  year={2023}
}

@article{wang2018dataset,
  title={Dataset distillation},
  author={Wang, Tongzhou and Zhu, Jun-Yan and Torralba, Antonio and Efros, Alexei A},
  journal={arXiv preprint arXiv:1811.10959},
  year={2018}
}

@article{settles2009active,
  title={Active learning literature survey},
  author={Settles, Burr},
  year={2009},
  publisher={University of Wisconsin-Madison Department of Computer Sciences}
}

@inproceedings{bengio2009curriculum,
  title={Curriculum learning},
  author={Bengio, Yoshua and Louradour, J{\'e}r{\^o}me and Collobert, Ronan and Weston, Jason},
  booktitle={Proceedings of the 26th annual international conference on machine learning},
  pages={41--48},
  year={2009}
}

@article{kumar2010self,
  title={Self-paced learning for latent variable models},
  author={Kumar, M and Packer, Benjamin and Koller, Daphne},
  journal={Advances in neural information processing systems},
  volume={23},
  year={2010}
}

@article{zhang2016understanding,
  title={Understanding deep learning requires rethinking generalization},
  author={Zhang, Chiyuan and Bengio, Samy and Hardt, Moritz and Recht, Benjamin and Vinyals, Oriol},
  journal={arXiv preprint arXiv:1611.03530},
  year={2016}
}

@inproceedings{bender2021dangers,
  title={On the dangers of stochastic parrots: Can language models be too big?🦜},
  author={Bender, Emily M and Gebru, Timnit and McMillan-Major, Angelina and Shmitchell, Shmargaret},
  booktitle={Proceedings of the 2021 ACM conference on fairness, accountability, and transparency},
  pages={610--623},
  year={2021}
}

@article{matiisen2019teacher,
  title={Teacher--student curriculum learning},
  author={Matiisen, Tambet and Oliver, Avital and Cohen, Taco and Schulman, John},
  journal={IEEE transactions on neural networks and learning systems},
  volume={31},
  number={9},
  pages={3732--3740},
  year={2019},
  publisher={IEEE}
}

@book{vygotsky1978mind,
  title={Mind in society: The development of higher psychological processes},
  author={Vygotsky, Lev S},
  volume={86},
  year={1978},
  publisher={Harvard university press}
}

@article{martinez2019item,
  title={Item response theory in AI: Analysing machine learning classifiers at the instance level},
  author={Mart{\'\i}nez-Plumed, Fernando and Prud{\^e}ncio, Ricardo BC and Mart{\'\i}nez-Us{\'o}, Adolfo and Hern{\'a}ndez-Orallo, Jos{\'e}},
  journal={Artificial intelligence},
  volume={271},
  pages={18--42},
  year={2019},
  publisher={Elsevier}
}

@inproceedings{wang2019dynamic,
  title={Dynamic curriculum learning for imbalanced data classification},
  author={Wang, Yiru and Gan, Weihao and Yang, Jie and Wu, Wei and Yan, Junjie},
  booktitle={Proceedings of the IEEE/CVF international conference on computer vision},
  pages={5017--5026},
  year={2019}
}

@article{lee2019estimating,
  title={Estimating student ability and problem difficulty using item response theory (IRT) and TrueSkill},
  author={Lee, Youngjin},
  journal={Information Discovery and Delivery},
  volume={47},
  number={2},
  pages={67--75},
  year={2019},
  publisher={Emerald Publishing Limited}
}

@article{chen2023alpagasus,
  title={Alpagasus: Training a better alpaca with fewer data},
  author={Chen, Lichang and Li, Shiyang and Yan, Jun and Wang, Hai and Gunaratna, Kalpa and Yadav, Vikas and Tang, Zheng and Srinivasan, Vijay and Zhou, Tianyi and Huang, Heng and others},
  journal={arXiv preprint arXiv:2307.08701},
  year={2023}
}

@inproceedings{li2024quantity,
  title={From quantity to quality: Boosting llm performance with self-guided data selection for instruction tuning},
  author={Li, Ming and Zhang, Yong and Li, Zhitao and Chen, Jiuhai and Chen, Lichang and Cheng, Ning and Wang, Jianzong and Zhou, Tianyi and Xiao, Jing},
  booktitle={Proceedings of the 2024 Conference of the North American Chapter of the Association for Computational Linguistics: Human Language Technologies (Volume 1: Long Papers)},
  pages={7602--7635},
  year={2024}
}

@article{yin2024compute,
  title={Compute-constrained data selection},
  author={Yin, Junjie Oscar and Rush, Alexander M},
  journal={arXiv preprint arXiv:2410.16208},
  year={2024}
}

@article{wang2025data,
  title={Data whisperer: Efficient data selection for task-specific llm fine-tuning via few-shot in-context learning},
  author={Wang, Shaobo and Jin, Xiangqi and Wang, Ziming and Wang, Jize and Zhang, Jiajun and Li, Kaixin and Wen, Zichen and Li, Zhong and He, Conghui and Hu, Xuming and others},
  journal={arXiv preprint arXiv:2505.12212},
  year={2025}
}

@article{cobbe2021training,
  title={Training verifiers to solve math word problems},
  author={Cobbe, Karl and Kosaraju, Vineet and Bavarian, Mohammad and Chen, Mark and Jun, Heewoo and Kaiser, Lukasz and Plappert, Matthias and Tworek, Jerry and Hilton, Jacob and Nakano, Reiichiro and others},
  journal={arXiv preprint arXiv:2110.14168},
  year={2021}
}

@article{jin2021disease,
  title={What disease does this patient have? a large-scale open domain question answering dataset from medical exams},
  author={Jin, Di and Pan, Eileen and Oufattole, Nassim and Weng, Wei-Hung and Fang, Hanyi and Szolovits, Peter},
  journal={Applied Sciences},
  volume={11},
  number={14},
  pages={6421},
  year={2021},
  publisher={MDPI}
}

@article{dubey2024llama,
  title={The llama 3 herd of models},
  author={Dubey, Abhimanyu and Jauhri, Abhinav and Pandey, Abhinav and Kadian, Abhishek and Al-Dahle, Ahmad and Letman, Aiesha and Mathur, Akhil and Schelten, Alan and Yang, Amy and Fan, Angela and others},
  journal={arXiv preprint arXiv:2407.21783},
  year={2024}
}

@article{yang2025qwen3,
  title={Qwen3 technical report},
  author={Yang, An and Li, Anfeng and Yang, Baosong and Zhang, Beichen and Hui, Binyuan and Zheng, Bo and Yu, Bowen and Gao, Chang and Huang, Chengen and Lv, Chenxu and others},
  journal={arXiv preprint arXiv:2505.09388},
  year={2025}
}

@article{jiang2023mistral,
  title={Mistral 7B},
  author={Jiang, Albert Q. and Sablayrolles, Alexandre and Roux, Antoine and Mensch, Arthur and others},
  journal={arXiv preprint arXiv:2310.06825},
  year={2023}
}

@article{gao2025principled,
  title={Principled data selection for alignment: The hidden risks of difficult examples},
  author={Gao, Chengqian and Li, Haonan and Liu, Liu and Xie, Zeke and Zhao, Peilin and Xu, Zhiqiang},
  journal={arXiv preprint arXiv:2502.09650},
  year={2025}
}

@inproceedings{zhou2025davir,
  title={Davir: Data selection via implicit reward for large language models},
  author={Zhou, Haotian and Liu, Tingkai and Ma, Qianli and Zhang, Yufeng and Yuan, Jianbo and Liu, Pengfei and You, Yang and Yang, Hongxia},
  booktitle={Proceedings of the 63rd Annual Meeting of the Association for Computational Linguistics (Volume 1: Long Papers)},
  pages={9220--9237},
  year={2025}
}

@article{li2025learnalign,
  title={Learnalign: Reasoning data selection for reinforcement learning in large language models based on improved gradient alignment},
  author={Li, Shipeng and Li, Shikun and Yang, Zhiqin and Zhang, Xinghua and Chen, Gaode and Xia, Xiaobo and Liu, Hengyu and Peng, Zhe},
  journal={arXiv preprint arXiv:2506.11480},
  year={2025}
}

@article{ming2025ideal,
  title={IDEAL: Data Equilibrium Adaptation for Multi-Capability Language Model Alignment},
  author={Ming, Chenlin and Qu, Chendi and Cai, Mengzhang and Pei, Qizhi and Pan, Zhuoshi and Li, Yu and Duan, Xiaoming and Wu, Lijun and He, Conghui},
  journal={arXiv preprint arXiv:2505.12762},
  year={2025}
}

@article{mekala2024smaller,
  title={Smaller language models are capable of selecting instruction-tuning training data for larger language models},
  author={Mekala, Dheeraj and Nguyen, Alex and Shang, Jingbo},
  journal={arXiv preprint arXiv:2402.10430},
  year={2024}
}

@article{ge2025capability,
  title={Capability Salience Vector: Fine-grained Alignment of Loss and Capabilities for Downstream Task Scaling Law},
  author={Ge, Qiming and Xing, Shuhao and Gao, Songyang and Zhou, Yunhua and Zou, Yicheng and Zhang, Songyang and Chen, Zhi and Yan, Hang and Zhang, Qi and Guo, Qipeng and others},
  journal={arXiv preprint arXiv:2506.13216},
  year={2025}
}

@inproceedings{xia2023training,
  title={Training trajectories of language models across scales},
  author={Xia, Mengzhou and Artetxe, Mikel and Zhou, Chunting and Lin, Xi Victoria and Pasunuru, Ramakanth and Chen, Danqi and Zettlemoyer, Luke and Stoyanov, Veselin},
  booktitle={Proceedings of the 61st Annual Meeting of the Association for Computational Linguistics (Volume 1: Long Papers)},
  pages={13711--13738},
  year={2023}
}

@inproceedings{suzgun2023challenging,
  title={Challenging big-bench tasks and whether chain-of-thought can solve them},
  author={Suzgun, Mirac and Scales, Nathan and Sch{\"a}rli, Nathanael and Gehrmann, Sebastian and Tay, Yi and Chung, Hyung Won and Chowdhery, Aakanksha and Le, Quoc and Chi, Ed and Zhou, Denny and others},
  booktitle={Findings of the Association for Computational Linguistics: ACL 2023},
  pages={13003--13051},
  year={2023}
}

@inproceedings{ye2023predictable,
  title={How predictable are large language model capabilities? a case study on big-bench},
  author={Ye, Qinyuan and Fu, Harvey and Ren, Xiang and Jia, Robin},
  booktitle={Findings of the Association for Computational Linguistics: EMNLP 2023},
  pages={7493--7517},
  year={2023}
}

\appendix
\clearpage
\onecolumn
\section{Appendix}

\subsection*{Appendix Overview}

This appendix includes the following materials:

\begin{itemize}
  \item \textbf{A.1 Figure 5: Difficulty distributions of ZPD-selected samples across different backbone models}
  \item \textbf{A.2 Experimental Settings}: Fine-tuning configurations for all models.
  \item \textbf{A.3 Dataset Details}: Descriptions of MedQA, GSM8K, and the AgriQA dataset.
  \item \textbf{A.4 Algorithm of ZPD-Detector}: Step-by-step pseudocode for our data selection pipeline.
  \item \textbf{A.5 Model Ability Estimation}: Binary search procedure under the Rasch model.
  \item \textbf{A.6 Task-Specific Prompt Design}: Prompt formats tailored to different task types.
  \item \textbf{A.7 AgriQA Data Synthesis Prompt}: Instruction for generating synthetic agriculture data.
\end{itemize}

\subsection{Capability-adaptive behavior of ZPD}

\begin{figure*}[t]
\centering
\includegraphics[width=1\linewidth]{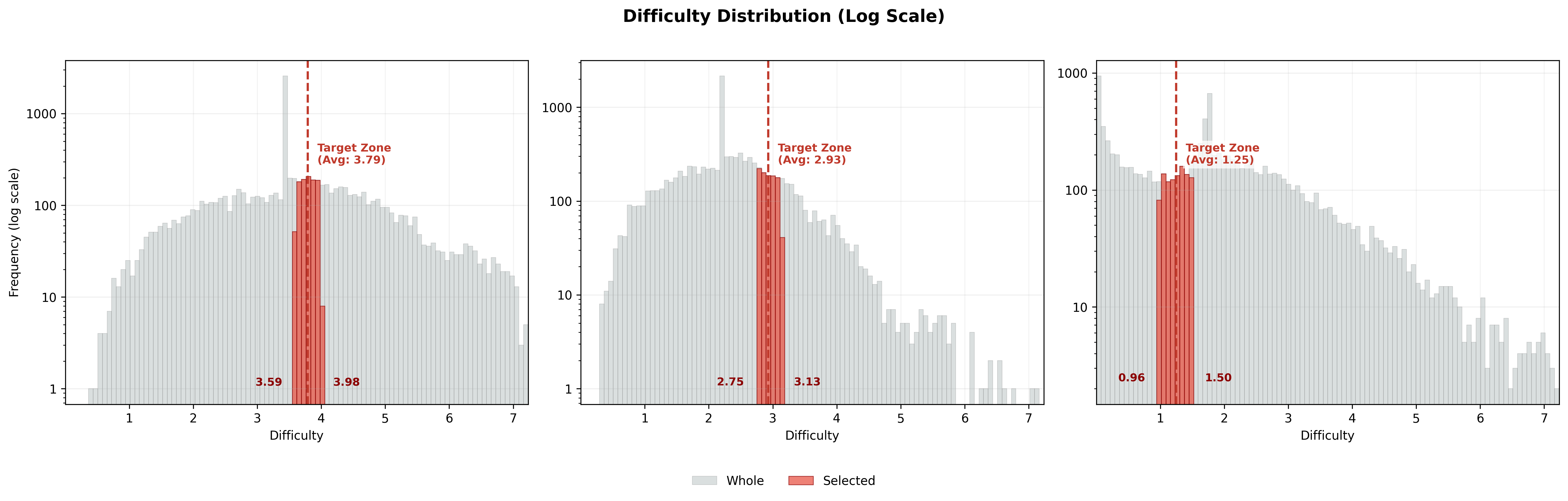}

\caption{
Difficulty distributions of ZPD-selected samples across different backbone models.
From left to right: LLaMA3-8B-Instruct, Qwen3-8B, and Mistral-7B-Instruct.
}
\label{fig:Grad1}
\end{figure*}

\subsection{Experimental Settings}

\textbf{OS:} Ubuntu 22.04.1 LTS / \textbf{GPUs:} 8 $\times$ NVIDIA GeForce RTX 4090  /  {CUDA (driver):} CUDA 12.6 / \textbf{Python:} 3.12.9 / \textbf{Conda version:} 25.1.1

\textbf{Notes}
\begin{itemize}
  \item All runs were performed under identical software and hardware settings.
\end{itemize}

\begin{table}[h]
\centering
\small
\begin{tabular}{lcccccc}
\toprule
\textbf{Model} & \textbf{LoRA} & \textbf{LR} & \textbf{Epochs} & \textbf{Batch Size} & \textbf{Scheduler} & \textbf{Framework} \\
\midrule
LLaMA3-8B-Instruct       & Applied & $2\mathrm{e}{-5}$ & 5 & 8 & Linear + Cosine & \texttt{ms-swift} \\
Mistral-7B-Instruct-v0.2 & Applied & $2\mathrm{e}{-5}$ & 5 & 8 & Linear + Cosine & \texttt{ms-swift} \\
Qwen3-8B                 & Applied & $5\mathrm{e}{-5}$ & 5 & 8 & Linear + Cosine & \texttt{ms-swift} \\
\bottomrule
\end{tabular}
\caption{Fine-tuning hyperparameters and configurations for all models. All experiments use LoRA-based parameter-efficient tuning, a unified batch size of 8, and the \texttt{ms-swift} training framework with linear warmup and cosine decay scheduling.}
\label{tab:implementation-details}
\end{table}

\subsection{Datasets Details}

\begin{tcolorbox}[
  title=\textbf{Datasets Used in Our Experiments},
  colback=gray!5!white,
  colframe=black!60!white,
  boxrule=0.8pt,
  sharp corners=south,
  breakable,
  enhanced
]

We evaluate our method on three datasets spanning diverse reasoning tasks and domains:

\begin{itemize}
  \item \textbf{MedQA}~\cite{jin2021disease}: A biomedical question-answering dataset composed of multiple-choice questions from professional medical licensing exams. The questions cover clinical reasoning, pathology, and general medical knowledge. We follow the official split: 10,178 training samples and 1,273 test samples.
  
  \item \textbf{GSM8K}~\cite{cobbe2021training}: A high-quality benchmark of approximately 8.5K linguistically diverse primary-school-level math word problems. Solving these problems typically requires multi-step arithmetic reasoning, making GSM8K a standard testbed for evaluating mathematical logical ability.
  
  \item \textbf{AgriQA (Synthetic)}: A domain-specific instruction dataset we construct to assess performance in the agricultural domain. It contains 9,955 samples (8,788 train / 1,167 test), generated using DeepSeek-V3 over curated agricultural corpora. The questions target real-world subfields such as pest management, basic agronomy, and agricultural technology dissemination. This dataset fills the gap where general-purpose LLMs often lack vertical domain competence.
\end{itemize}

\end{tcolorbox}

\vspace{-2.5cm}
\subsection{Algorithm of ZPD-Detector}

\renewcommand{\baselinestretch}{1.15}

\begin{algorithm}[H]
\caption{\textbf{ZPD-Detector for Data Selection}}
\label{alg:zpd-detector}
\begin{flushleft}
\textbf{Input:} \\
\quad $\mathcal{M}$ — Pre-trained (or current) language model \\
\quad $\mathcal{D}$ — Full instruction dataset $\mathcal{D} = \{(x_i, y_i)\}_{i=1}^N$ \\
\quad $f$ — Token-level NLL loss \\
\quad $\text{Eval}(\cdot)$ — Correctness function returning $r_i \in \{0,1\}$ \\
\quad $\rho$ — Budget ratio $(0 < \rho \leq 1)$

\vspace{0.5em}
\textbf{Output:} \\
\quad Selected subset $\mathcal{D}' \subseteq \mathcal{D}$ of size $|\mathcal{D}'| = \lceil \rho \cdot |\mathcal{D}| \rceil$
\end{flushleft}

\begin{algorithmic}[1]
\State \textbf{Step 1:} Compute raw difficulty via model loss
\For{each $(x_i, y_i) \in \mathcal{D}$}
    \State $RawDiff_i \gets -\frac{1}{L_i} \sum_{t=1}^{L_i} \log P(y_{i,t} \mid y_{i,<t}, x_i; \mathcal{M})$
\EndFor
\State $\mu \gets \text{Mean}(RawDiff)$ \Comment{global mean difficulty}
\Statex \algrule

\State \textbf{Step 2:} Calibrate difficulty with correctness
\For{each $(x_i, y_i) \in \mathcal{D}$}
    \State $\hat{y}_i \gets \text{Generate}(\mathcal{M}, x_i)$ \Comment{greedy or task decoding}
    \State $r_i \gets \text{Eval}(\hat{y}_i, y_i)$
    \State $b_i \gets RawDiff_i + (1 - r_i) \cdot \max(0, \mu - RawDiff_i)$ \Comment{calibrated difficulty}
\EndFor
\Statex \algrule

\State \textbf{Step 3:} Estimate model ability $\theta$ via Rasch (1PL IRT)
\State Initialize $\theta \gets 0$
\State Maximize log-likelihood:
\[
\ell(\theta) = \sum_{i=1}^{N} \left[ r_i \log \sigma(\theta - b_i) + (1 - r_i) \log (1 - \sigma(\theta - b_i)) \right]
\]
\State $\hat{\theta} \gets \arg\max_\theta \ell(\theta)$
\Statex \algrule

\State \textbf{Step 4:} Compute ZPD proximity scores
\For{each $(x_i, y_i) \in \mathcal{D}$}
    \State $p_i \gets \sigma(\hat{\theta} - b_i)$ \Comment{predicted correctness}
    \State $s_i \gets p_i (1 - p_i)$ \Comment{ZPDScore}
\EndFor
\Statex \algrule

\State \textbf{Step 5:} Select top samples under budget
\State $k \gets \lceil \rho \cdot |\mathcal{D}| \rceil$
\State $\mathcal{D}' \gets \text{Top-}k(\mathcal{D}, \text{scores}=s_i)$
\State \Return $\mathcal{D}'$
\end{algorithmic}
\end{algorithm}

\subsection{Details of Model Ability Estimation via Rasch Model}

\begin{tcolorbox}[
  title=\textbf{Model Ability Estimation via Rasch Model},
  colback=gray!5!white,
  colframe=black!60!white,
  boxrule=0.8pt,
  sharp corners=south,
  breakable,
  enhanced
]

To estimate the model’s current ability from its binary response outcomes, we adopt the Rasch model (1PL IRT) and solve the following maximum likelihood estimation (MLE) objective:

\[
\hat{\theta} = \arg\max_\theta \sum_{i=1}^{N} \left[ r_i \log \sigma(\theta - b_i) + (1 - r_i) \log (1 - \sigma(\theta - b_i)) \right]
\]

Here, \( r_i \in \{0, 1\} \) indicates whether the model correctly answered item \(i\), \( b_i \) is the item's calibrated difficulty, and \( \sigma(\cdot) \) is the logistic function.

\medskip
We apply a binary search procedure to estimate \(\hat{\theta}\), operating within a dynamic interval:

\[
\theta \in [\min(b_i) - 30,\ \max(b_i) + 30]
\]

The margin of 30 is chosen to be sufficiently large such that the logistic function is numerically saturated (i.e., $\sigma(\pm30) \approx 1$ or $0$), ensuring that the search bound safely brackets the optimum without affecting the MLE solution.

\textbf{Stopping Criteria:}
\begin{itemize}
  \item The gap between predicted and actual number of correct responses is within 1 sample.
  \item The update difference in \(\theta\) between iterations is less than \(10^{-6}\).
\end{itemize}

\medskip
This procedure yields a stable and efficient estimate of model ability \(\hat{\theta}\) while keeping the item difficulties \(\{b_i\}\) fixed.
\end{tcolorbox}

\subsection{Task-Specific Prompt Design}

\begin{tcolorbox}[
  title=\textbf{Task-Specific Prompt Design},
  colback=gray!5!white,
  colframe=black!60!white,
  boxrule=0.8pt,
  sharp corners=south,
  breakable,
  enhanced
]

\textbf{For MedQA and AgriQA: Accuracy-Priority Prompting}

To emphasize factual correctness in domain-specific question answering, we apply the following instruction format for medical and agricultural tasks:

\begin{quote}
Please carefully read the question and all provided options. Select the most accurate answer according to your knowledge, and justify your choice if necessary. Accuracy takes precedence over fluency or politeness.
\end{quote}

\medskip
\textbf{For GSM8K: Format-Constrained Math Reasoning}

For math word problems in the GSM8K dataset, we adopt strict output formatting to ensure evaluation clarity:

\begin{quote}
Please include intermediate reasoning steps in your answer. The final numeric answer must appear \textbf{only} within a box using the format: \texttt{\#\#\#\# <number>}.\\
Do not provide any additional explanation or commentary.
\end{quote}

\textbf{Example output:}
\begin{lstlisting}
The total number of apples is:
3 + 5 = 8
#### 8
\end{lstlisting}

\end{tcolorbox}

\subsection{Data Synthesis Prompt for AgriQA}
\begin{tcolorbox}[
  title=\textbf{Data Synthesis Prompt for AgriQA},
  colback=gray!5!white,
  colframe=black!60!white,
  boxrule=0.8pt,
  sharp corners=south,
  breakable,
  enhanced
]

You are an AI assistant and an expert in the field of agriculture, tasked with generating detailed and accurate multiple-choice questions based on the provided reference material.

You will be given a content reference. Based on that, generate a specified number of \textbf{AlpacaItems}, each consisting of:
\begin{itemize}
  \item an instruction (i.e., a multiple-choice question),
  \item an empty input field,
  \item and a correct answer, which must be a single letter (A, B, C, or D).
\end{itemize}

Guidelines:
\begin{itemize}
  \item Each instruction must be directly answerable only by referring to the provided content.
  \item The context must be agriculture-related.
  \item Add a numeric prefix to each item to indicate its position.
  \item Avoid questions answerable via general commonsense knowledge.
  \item Ensure broad coverage of the most important points in the reference content.
  \item Do not include the imagined real-world scenario, but use it internally to inform realistic and diverse questions.
\end{itemize}

The format of each AlpacaItem must follow the instruction-input-output schema, with the \texttt{input} field left blank.

\textbf{Input Parameters:}
\begin{itemize}
  \item \texttt{\{content\}}: the reference agricultural material,
  \item \texttt{\{examples\_str\}}: optional examples for in-context learning,
  \item \texttt{\{n\_items\}}: number of AlpacaItems to generate,
  \item \texttt{\{start\_num\}}: the starting index number.
\end{itemize}

\textbf{Expected Output:} A sequence of exactly \texttt{n\_items} labeled multiple-choice questions with answers from \texttt{A--D}.

\end{tcolorbox}

\end{document}